\pdfoutput=1

\documentclass[11pt]{article}

\usepackage{ACL2023}

\usepackage{times}
\usepackage{latexsym}

\usepackage[T1]{fontenc}

\usepackage[utf8]{inputenc}

\usepackage{microtype}

\usepackage{inconsolata}

\usepackage{graphicx}
\usepackage{booktabs}
\usepackage{amsmath,amssymb,bm}
\usepackage{tabularx}
\usepackage{here}
\usepackage{multicol}

\title{SciReviewGen: A Large-scale Dataset for Automatic Literature Review Generation}

\author{Tetsu Kasanishi$^{1}$ \quad \textbf{Masaru Isonuma}$^{1}$ \quad \textbf{Junichiro Mori}$^{1, 2}$ \quad \textbf{Ichiro Sakata}$^{1}$ \bigskip\\
		$^1$ The University of Tokyo \quad
		$^2$ RIKEN Center for Advanced Intelligence Project \\
  		\texttt{\{kasanishi, isonuma, isakata\}@ipr-ctr.t.u-tokyo.ac.jp} \\ 
  		\texttt{mori@mi.u-tokyo.ac.jp} \qquad
  		}

\begin{document}
\maketitle
\begin{abstract}
Automatic literature review generation is one of the most challenging tasks in natural language processing.
Although large language models have tackled literature review generation, the absence of large-scale datasets has been a stumbling block to the progress.
We release \emph{SciReviewGen}, consisting of over 10,000 literature reviews and 690,000 papers cited in the reviews.
Based on the dataset, we evaluate recent transformer-based summarization models on the literature review generation task, including Fusion-in-Decoder \cite{izacard-grave-2021-leveraging} extended for literature review generation.
Human evaluation results show that some machine-generated summaries are comparable to human-written reviews, while revealing the challenges of automatic literature review generation such as hallucinations and a lack of detailed information.
Our dataset and code are available at \url{https://github.com/tetsu9923/SciReviewGen}.
\end{abstract}

\section{Introduction}
Scientific document processing has been a topic of interest in the frontiers of natural language processing (NLP) \cite{cohan-etal-2022-overview}.
Although neural-based NLP models have achieved remarkable success in diverse areas, scientific documents present distinct challenges, such as longer inputs, technical terms, and complex logic.
These challenges have motivated NLP researchers to undertake various studies on scientific documents, such as scientific document summarization, retrieval, and information extraction \cite{cohan-etal-2018-discourse, beltagy-etal-2019-scibert, cohan-etal-2020-specter, yuan2022can}.

\begin{figure}[t]
 \begin{center}
  \includegraphics[scale=0.135]{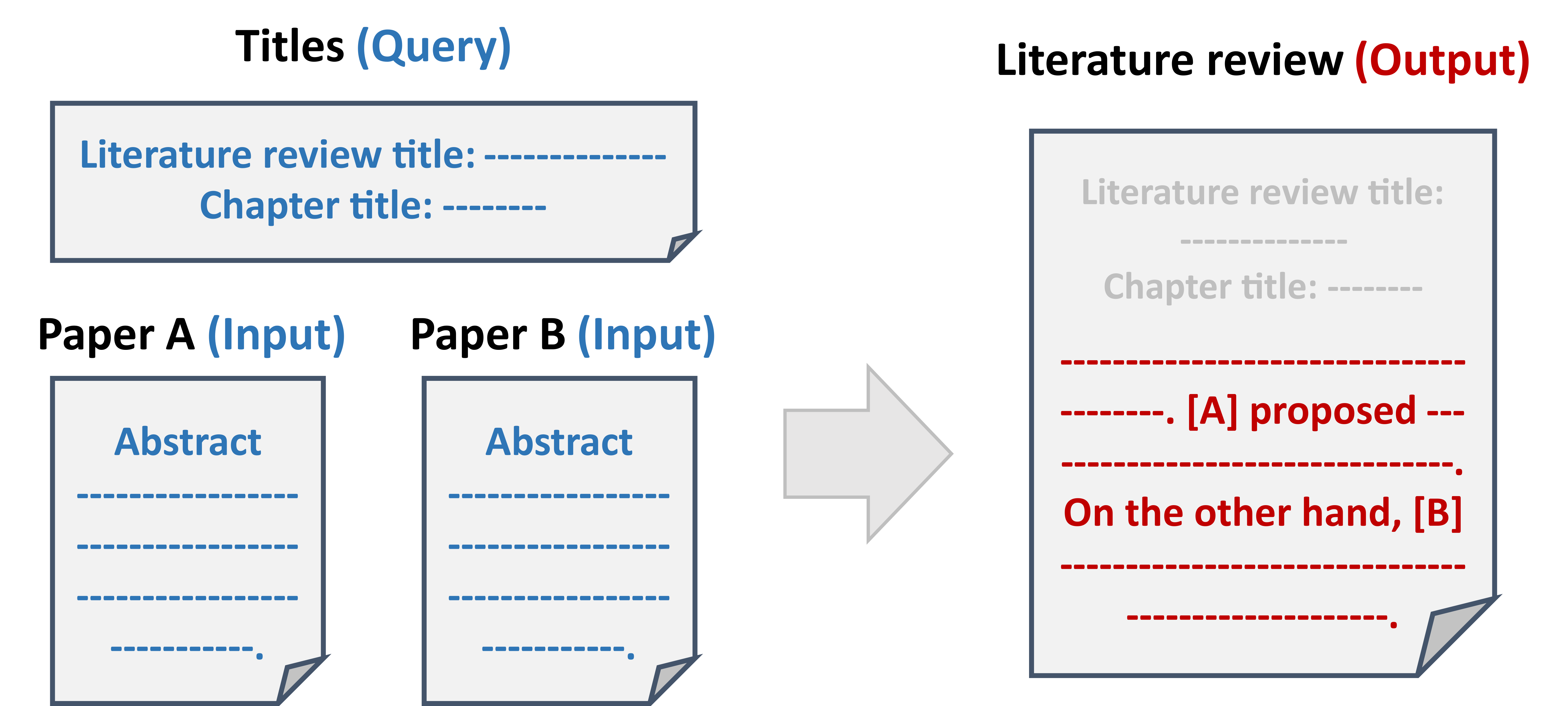}
  \caption{An overview of the literature review generation task in SciReviewGen.}
  \label{fig:overview}
  \vspace{-0.5\baselineskip}
 \end{center}
\end{figure}

Automatic literature review generation is one of the most attractive research topics in scientific document processing.
A literature review is a summary of scientific papers written by experts to comprehend previous findings \cite{jaidka-etal-2013-deconstructing}.
\citet{bornmann2015growth} investigated that the number of published scientific papers doubles every nine years, increasing the demand for literature reviews in diverse research areas.
Automatic literature review generation significantly benefits researchers by expanding their studies into new research fields.

However, only a few studies have addressed automatic literature review generation.
For example, \citet{taylor2022galactica} recently proposed GALACTICA, a large-scale language model trained on 48 million scientific papers.
GALACTICA was made publicly available to demonstrate its ability to generate literature reviews; however, it was shut down within a few days owing to the hallucination problem \cite{Galactic18:online}.
As there are no large-scale literature review datasets, applying data-hungry supervised neural summarization models is difficult. 
The absence of large-scale datasets is a significant bottleneck in research on automatic literature review generation.

In this study, we pioneer the research of automatic literature review generation by providing a large-scale dataset based on the Semantic Scholar Open Research Corpus \cite[S2ORC;][]{lo-etal-2020-s2orc}.
We release \emph{SciReviewGen}, which consists of over 10,000 literature reviews in the field of computer science and 690,000 papers cited in the reviews. 
As our dataset is created in a domain-agnostic way, it is possible to create datasets in other scientific fields, such as medical and biological sciences.

Figure \ref{fig:overview} shows an overview of the literature review generation task.
We regard it as a query-focused multi-document summarization (MDS) task.
The inputs are the abstracts of papers cited in the reviews, and the queries are the titles of the reviews and chapters, which specify the topics in the reviews.
In the actual writing process of a literature review, we need to decide on the papers to cite in the review and group them into several chapters.
As the first step for automatic literature review generation, we exclude those processes from our scope and focus on summarization given the cited papers and chapter division.
As SciReviewGen and S2ORC include bibliographic information on reviews and their cited papers (e.g., DOI, citation, and chapter division), our dataset can be used for end-to-end literature review generation.

Based on our dataset, we evaluate recent transformer-based summarization models for the literature review generation task.
As current summarization models cannot simultaneously generate the entire text of reviews, we split a review into chapters and evaluate each of the generated chapters.
In addition to recent models, such as Big Bird \cite{zaheer2020big} and Fusion-in-Decoder \cite[FiD;][]{izacard-grave-2021-leveraging}, we propose Query-weighted Fusion-in-Decoder (QFiD), a simple extension of FiD for query-focused MDS. 
As shown in the experimental results, our proposed model outperforms the other models by focusing on the contents concerning the query.

Finally, we conduct a human evaluation of the generated reviews and compare them with human-written reviews.
The human evaluation results clarified that we have not reached the fully automatic literature review generation stage due to issues such as hallucinations and less informativeness.
However, we obtained promising results, showing that approximately 30\% of the generated chapters are competitive or superior to human-written reviews.
Our dataset and evaluation results provide a basis for future research on automatic literature review generation.

\section{Related Work}
\subsection{Datasets for Scientific Document Summarization}
The most common datasets for document summarization are based on news articles, such as CNN/Daily Mail \cite{nallapati-etal-2016-abstractive}, XSum \cite{narayan-etal-2018-dont}, and Multi-News \cite{fabbri-etal-2019-multi}.
On the other hand, there are many datasets for scientific document summarization.
\citet{cohan-etal-2018-discourse} released arXiv and PubMed datasets, commonly used for abstract generation tasks.
\citet{lu-etal-2020-multi-xscience} proposed Multi-XScience, which aims to generate a related work section by using the abstract of a subject paper and papers cited in its related work section.
While related work section generally describes the position of the subject paper w.r.t. the previous studies, literature reviews generally provide the comprehensive summary of a research field.
Furtheremore, the length of input/output text of SciReviewGen is significantly longer than that of Multi-XScience (see Section \ref{statistics}).
Hence, our dataset has distinct challenges from Multi-XScience.

\citet{deyoung-etal-2021-ms} proposed MS\^{}2 for the automatic generation of systematic reviews in biomedical science.
Systematic reviews integrate findings from all relevant studies to answer clearly formulated questions, such as the safety of public water fluoridation \cite{khan2003five}.
In contrast, literature reviews include various topics, such as the motivations behind the research topic, technical details of the methods, and their real-world applications.
Furthermore, the target summaries in  MS\^{}2 are very short and are written under an explicit methodology \cite{khan2003five}.
In contrast, literature reviews are significantly longer, and the writing style varies according to the author \cite{jaidka-etal-2013-deconstructing, jaidka2013literature}.
Therefore, SciReviewGen is more challenging than MS\^{}2 in terms of output diversity.

\subsection{Automatic Literature Review Generation}
Few studies have addressed the automatic generation of literature reviews.
For example, \citet{mohammad-etal-2009-using} applied unsupervised summarization methods, such as LexRank \cite{erkan2004lexrank}, to generate technical surveys of scientific papers.
\citet{agarwal-etal-2011-towards} proposed clustering-based extractive methods for generating summaries of co-cited papers.
However, these methods do not aim to generate literature reviews, and only a few dozen gold summaries are used for evaluation instead of existing literature reviews.
While \citet{jaidka-etal-2013-deconstructing} claimed that they conducted literature review generation, no technical details of the model are described.
In contrast to these studies, we first release a large-scale dataset for literature review generation and intensively evaluate recent models by both automatic and human evaluation.

\subsection{Transformer-based long document / query-focused summarization}
Transformers \cite{vaswani2017attention} have shown remarkable success in document summarization.
Standard Transformer-based models can accept up to only 512-1024 tokens at once due to the high computational cost of the self-attention mechanism.
Recently, various methods have been proposed to overcome this limitation \cite{beltagy2020longformer}, such as the sparse attention mechanism used in Big Bird \cite{zaheer2020big}.
FiD \cite{izacard-grave-2021-leveraging} is a Transformer encoder-decoder model that allows multiple documents to be input.
Although initially designed for open-domain question answering, it can be applied to MDS tasks \cite{deyoung-etal-2021-ms, vig-etal-2022-exploring}.

Query-focused summarization (QFS) aims to generate summaries related to user-specified queries \cite{vig-etal-2022-exploring}.
Recent studies have applied Transformers to QFS, but most of them simply concatenate queries into input documents \cite{vig-etal-2022-exploring, laskar-etal-2022-domain}.

As mentioned in Section \ref{dataset}, SciReviewGen has an average input length longer than 1024 tokens and contains the titles of literature reviews and chapters as queries.
Therefore, we extend FiD for query-focused summarization to tackle the task of literature review generation.
Our proposed QFiD explicitly considers the relevance of each input document to the queries.

\section{Task Definition \& Dataset\label{dataset}}
We now describe the literature review generation task and the SciReviewGen dataset, which is created using S2ORC.
The data collection process and dataset statistics are presented below.

\subsection{Task Definition\label{definition}}
As there are no previous datasets for literature review generation, we first describe the definition of the literature review generation task.

\paragraph{Target Text}
Ideally, the entire text of a literature review should be used as target.
However, as current summarization models can generate relatively short summaries of less than a thousand tokens \cite{fabbri-etal-2019-multi, narayan-etal-2018-dont}, it is difficult to generate the entire text of literature reviews simultaneously.
Therefore, we split a review paper into chapters in the following experiments and use each chapter as a target text.
In addition, as each chapter of the literature review generally discusses different topics, we assume that each chapter can be generated independently as the first step for automatic literature review generation.

\paragraph{Input Text}
The following data are input for the literature review generation task:
\emph{abstracts of cited papers}, \emph{titles of literature reviews}, and \emph{titles of chapters}.
Here, \emph{cited papers} refer to those cited in each chapter.
The abstracts of the cited papers are used as the primary sources for the contents of the generated chapter.
Although it is desirable to input the full text of the cited papers, we use only abstracts, as approximately 30\% of them do not have access to the full text in the S2ORC dataset.
The titles of the review and chapter serve as queries. 
They suggest the topics described in each chapter.

\paragraph{Additional Inputs}
As SciReviewGen contains citation information, such as citation sentences and citation networks, they can be used as information sources that complement abstracts.
The citation sentences provide the cited paper's actual impact on the research community \cite{yasunaga2019scisummnet}, whereas citation networks provide the relationships between the cited papers. 
Furthermore, SciReviewGen are linked to S2ORC by $\mathrm{paper\_id}$.
Therefore, various metadata in S2ORC (e.g., DOI, journal, and semantic scholar URL) attached to the literature review and cited papers can be accessed.

\begin{table*}[t]
\centering
\scalebox{0.79}{
\begin{tabular}{lrrrrrrrr}
\toprule
dataset & train/valid/test & input len. & target len. & \# inputs & unigrams & bigrams & trigrams & 4-grams \\
\midrule
Multi-News & 44,972/5,622/5,622 & 2,103 & 264 & 2.79 & 16.87\% & 55.57\% & 74.44\% & 81.23\% \\
MS\^{}2 & 14,188/2,021/1,667 & 6,930 & 61 & 22.80 & 15.24\% & 62.35\% & 87.23\% & 95.27\% \\
Multi-XScience & 30,369/5,066/5,093 & 778 & 116 & 4.42 & 35.28\% & 81.57\% & 94.88\% & 97.89\% \\
SciReviewGen (original) & 9,187/484/459 & 12,503 & 8,082 & 68.00 & 17.88\% & 64.86\% & 90.56\% & 97.20\% \\
SciReviewGen (split) & 84,705/4,410/4,457 & 1,274 & 604 & 7.01 & 32.74\% & 80.23\% & 95.16\% & 98.09\% \\
\bottomrule
\end{tabular}
}
\caption{Comparison of large-scale multi-document summarization (MDS) datasets. The number of target summaries in each split (train/valid/test), the average number of input tokens per target summaries (input len.), the average number of target tokens (target len.), the average number of input documents per target summary (\# inputs), and percentage of novel n-grams are shown. SciReviewGen (original) sets the whole text of literature reviews as target, while SciReviewGen (split) sets each filtered chapter as target. We use SciReviewGen (split) in our experiments.}
\label{fig:stats}
\end{table*}

\subsection{Dataset Construction \label{construction}}
We constructed SciReviewGen based on S2ORC \cite{lo-etal-2020-s2orc}, a large corpus of English academic papers.
First, as candidates for literature reviews, we extracted papers with access to full-text data where the field of study includes ``Computer Science,'' and the title contains either ``survey,'' 
 ``overview,'' ``literature review,'' or ``a review.''
This yielded 13,984 candidates for the literature reviews.

As the above candidates still contain many papers unrelated to literature reviews, we trained a SciBERT-based classifier \cite{beltagy-etal-2019-scibert} to extract appropriate literature reviews from the candidates.
We first created a gold-standard dataset of literature reviews to train the classifier.
We asked three annotators with computer science research backgrounds to annotate whether each candidate paper was suitable as a literature review following the two criteria:
1) Reviewing multiple scientific papers. Not reviewing general tools or books and not explaining a specific project or shared task;
2) Only reviewing scientific papers. Not proposing new methods, re-testing previous studies, or conducting questionnaires (i.e., the paper does not contain contents that cannot be generated only by the cited papers' information).

The above criteria were set so that the annotators could judge only from the title and abstract of a candidate paper.
They classified whether each paper was suitable as a literature review, and the class in which most annotators voted was used as the final annotation result.
The annotators classified 583 of 889 candidate papers as suitable and 306 as unsuitable, resulting in Cohen's kappa = 0.66.

The annotated papers were then split into a train/valid/test set containing 589/150/150 papers for training the SciBERT-based classifier.
Using the train/valid split, we fine-tuned the SciBERT classifier, which achieved $\textrm{precision} = 88\%$, $\textrm{recall} = 97\%$, and $\textrm{f1} = 92\%$ on the test split.
Using this classifier, we extracted 10,269 papers from 13,984 candidate papers, including 210,049 chapters and 698,049 cited papers.
As a result, we constructed SciReviewGen (original), consisting of the entire text of literature reviews, the titles of literature reviews and chapters, and the abstracts of the cited papers.

For our experiments, we split the literature reviews into chapters and excluded chapters that had access to less than two abstracts of their cited papers, leaving 93,572 chapters.
This split version is denoted as SciReviewGen (split).
As S2ORC does not contain the data of some cited papers, the number of filtered chapters will increase if we obtain the data of all cited papers.

Finally, to ensure that the test set includes only suitable papers, we set the human-annotated papers as the test sets and created the train/valid sets by randomly splitting the rest for both original and split version.
Furthermore, we removed the chapters in the test sets that have more than 20\% overlap of cited papers with one or more literature reviews in the training set.

\subsection{Dataset Statistics\label{statistics}}
Table \ref{fig:stats} presents the statistics of SciReviewGen compared with current large-scale MDS datasets, including Multi-News \cite{fabbri-etal-2019-multi}, MS\^{}2 \cite{deyoung-etal-2021-ms}, and Multi-XScience \cite{lu-etal-2020-multi-xscience}.
Regarding the split version, SciReviewGen has more than approximately twice as many summaries as the other datasets, which is more suitable for data-driven neural-based summarization models.
The target length is more than twice that of the other datasets.
SciReviewGen also has more input documents and a longer input length than Multi-XScience.
Furthermore, the original version presents distinct characteristics, such as significantly longer input/target text and more input documents than the others.
These characteristics would be the challenge for further research in automatic literature review generation.
Note that the ratio of input length to target length are relatively small in both versions; however, inputs can be complemented by additional information, such as body text and citation sentences.

Table \ref{fig:stats} also lists the percentage of novel n-grams in the target summary that do not appear in the input documents.
The target summaries in SciReviewGen contain more novel n-grams than those in Multi-News and MS\^{}2, indicating that SciReviewGen is more challenging and suitable for abstract summarization.
It is reasonable that both SciReviewGen (split) and Multi-XScience contain many novel n-grams because both the literature reviews and related work sections contain high-level summaries of the cited papers \cite{jaidka-etal-2013-deconstructing, jaidka2019characterizing}.

\section{Experiments\label{exp}}
We study the performance of the current document summarization models on the split version of SciReviewGen (hereinafter refered to as SciReviewGen).
We use the abstracts of the cited papers, the literature review titles, and the chapter titles as inputs.
As mentioned in Section \ref{statistics}, SciReviewGen has an average input length of longer than 1024 tokens and contains many novel n-grams.
In addition, it contains literature review titles and chapter titles that can be used as summarization queries.
Therefore, we employ query-focused abstractive summarization models that can accept long sequences for the literature review generation task.
We first experiment with several transformer-based models that simply concatenate queries into documents as encoder inputs.
We then propose the Query-weighted Fusion-in-Decoder (QFiD) that extends Fusion-in-Decoder (FiD) to explicitly consider each paper's relevance to queries. 

\subsection{Baseline Methods}

\begin{figure*}[t]
 \begin{center}
  \includegraphics[scale=0.085]{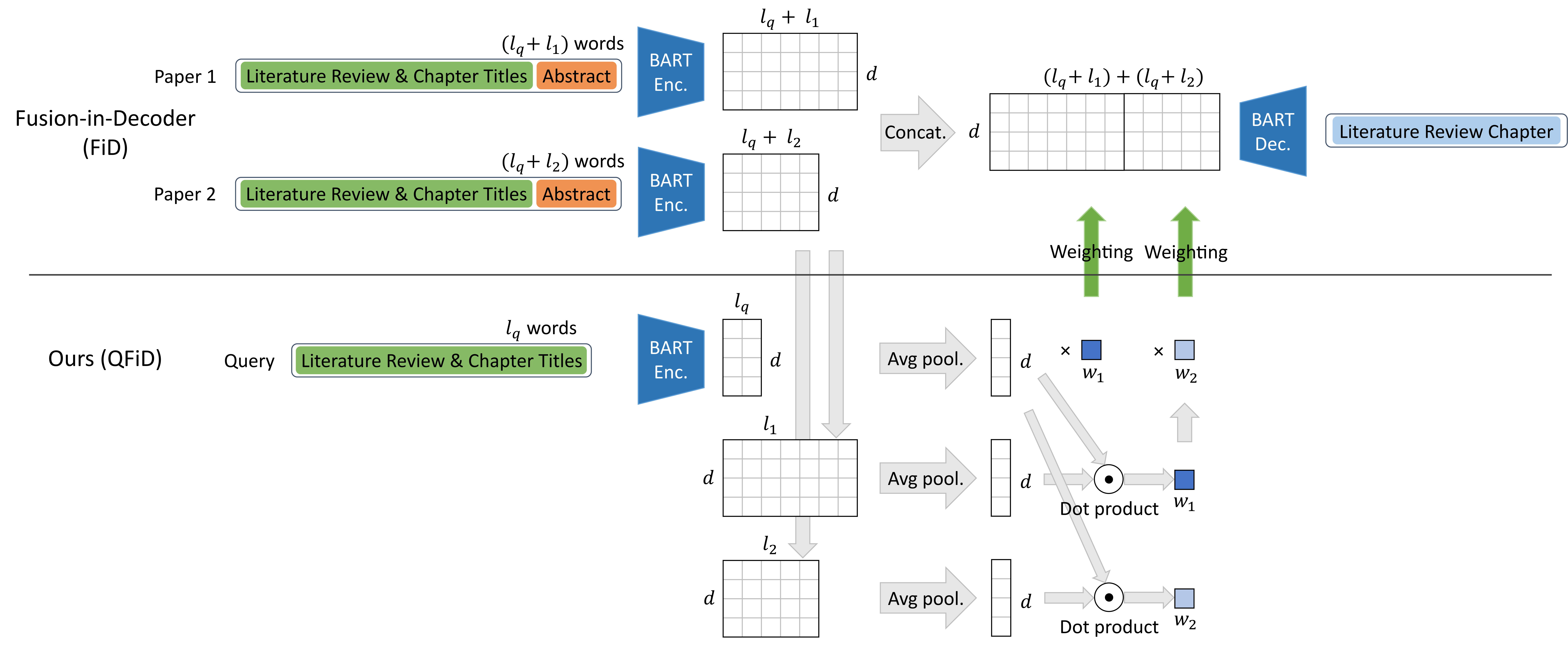}
  \caption{An overview of Fusion-in-Decoder (FiD) \cite{izacard-grave-2021-leveraging} and our Query-weighted Fusion-in-Decoder (QFiD) model, which extends FiD for query-focused MDS.}
  \label{fig:fid}
 \end{center}
\end{figure*}

We use LEAD, LexRank \cite{erkan2004lexrank}, ext-oracle, Big Bird \cite{zaheer2020big}, and FiD \cite{izacard-grave-2021-leveraging} as the baseline methods.

LEAD-k selects the first $k$ sentences from each input document and concatenates them as a summary.
LexRank is a graph-based unsupervised extractive method that considers a graph in which the sentences are nodes, and the similarities between the sentences are edges.
It calculates the importance of sentences using PageRank algorithm \cite{page1998pagerank} and extracts the top $l$ sentences with high importance as a summary.
Ext-oracle greedily selects $l$ sentences that maximize the ROUGE-2 scores between the selected sentences and the target summary.
Its results show the upper bound of an extractive system on SciReviewGen.
We set $k\!=\!1$ and $l\!=\!5$ such that the average summary length is the same as that of the abstractive models.

Big Bird simplifies the self-attention computation in the Transformer using the sparse attention mechanism, supporting longer inputs of up to approximately 16K tokens.
In our experiments, we use the model that was fine-tuned for summarization on arXiv dataset \cite{cohan-etal-2018-discourse}\footnote{\url{https://huggingface.co/google/bigbird-pegasus-large-arxiv}}.
We further fine-tuned it on SciReviewGen.
FiD is a Transformer encoder-decoder model that allows multiple documents to be input.
As shown in the upper part of Figure \ref{fig:fid}, FiD separately encodes multiple documents and concatenates their hidden states.
The hidden states are then input into the decoder together, which enables multiple documents to be simultaneously processed while capturing the relations among documents.
In our experiment, we initialized the weights of FiD using the BART-Large model \cite{lewis-etal-2020-bart} fine-tuned for summarization on the CNN/Daily Mail dataset\footnote{\url{https://huggingface.co/facebook/bart-large-cnn}}, and further fine-tuned it on SciReviewGen.

Note that we also evaluated the performance of GPT-3 model davinci \cite{brown2020language} on SciReviewGen with the prompt ``Summarize the above scientific papers focusing on the title and chapter title.''
However, it yielded almost no meaningful sentences, resulting in significantly lower ROUGE scores (ROUGE-1 = 9.77, ROUGE-2 = 1.25, ROUGE-L = 8.67).

\subsection{Query-weighted Fusion-in-Decoder (QFiD)}
This section describes our QFiD model that extends FiD to explicitly consider each paper's relevance to the queries.
As mentioned in Section \ref{definition}, the titles and chapter titles serve as queries that suggest the topic in each chapter.
The baseline methods simply concatenate these queries with the abstract of each cited paper.
For FiD, this simple approach makes the encoder consider the local relation between the queries and the words of each abstract.
However, the model cannot explicitly identify which cited papers are related to the queries.
In the literature review generation task, not all cited papers are related to a chapter's topic.
For example, when the chapter describes machine learning methods, it typically cites papers that describe datasets or evaluation metrics along with experimental results.
However, these papers are not directly related to the methods, and their contents should be less focused on.

For the aforementioned reason, we improved FiD to explicitly consider the relevance of each cited paper to the queries. 
Our model weights each cited paper according to its similarity to the query to identify which papers are more related to the topic in the chapter.
Specifically, as shown in the lower part of Figure \ref{fig:fid}, 
let $n$ be the number of the cited papers, $\bm{r}_m$ be the input token sequence of the $m$-th cited paper, and ${l_m}$ be its length for $m \!\in\! \{1, \ldots, n\}$.
Let $\bm{q}$ be the query that concatenates the title and chapter title, and $l_q$ be its length.
The hidden states of the $m$-th cited paper $H_m \in \mathbb{R}^{d \times \left(l_q + l_m\right)}$ and query $H_q \in \mathbb{R}^{d \times l_q}$ are obtained as follows:
\begin{eqnarray}
H_m &=& \mathrm{Enc}\left(\bm{q} + \bm{r}_m\right) \\
H_q &=& \mathrm{Enc}\left(\bm{q}\right)
\end{eqnarray}
where $\mathrm{Enc}$ is the BART encoder, and $d$ denotes the dimension of each hidden state.
Then, the feature vectors of the $m$-th cited paper $\bm{h}_m \in \mathbb{R}^{d}$ and query $\bm{h}_q \in \mathbb{R}^{d}$ are obtained as follows:
\begin{eqnarray}
\bm{h}_m &=& \mathrm{Avgpool}\left(H_m\right) \\
\bm{h}_q &=& \mathrm{Avgpool}\left(H_q\right)
\end{eqnarray}
where $\mathrm{Avgpool}$ is the operation for computing the average of the hidden states.
The similarity between the query and the $m$-th cited paper $w_m \in \mathbb{R}$ is obtained as the inner product of these vectors.
Subsequently, the hidden states of the $m$-th cited paper are weighted by $w_m$ and input to the BART decoder.
\begin{eqnarray}
w_m = 1 + \frac{\exp(\bm{h}_m^{\top} \bm{h}_q)}{\sum_{m=1}^{n} \exp(\bm{h}_m^{\top} \bm{h}_q)} \\
\bm{c} \sim \mathrm{Dec}\left(\left[w_1H_1; ... ; w_nH_n\right]\right)
\end{eqnarray}
where $\left[w_1H_1; ... ; w_nH_n\right]$ is the concatenation of matrices.
$\mathrm{Dec}$ denotes the BART decoder, and $\bm{c}$ is the generated chapter of the literature review.

\begin{table}[t]
\centering
\begin{tabularx}{\columnwidth}{|X|}
\hline
\small
Literature review title <s> Chapter title <s> Abstract of paper 1 <s> BIB001 </s> Literature review title <s> Chapter title <s> Abstract of paper 2 <s> BIB002 </s> ... </s> Literature review title <s> Chapter title <s> Abstract of paper N <s> BIB00N
\\ \hline
\end{tabularx}
\caption{Input data format.}
\label{fig:format}
\end{table}

\subsection{Implementation Details}
The input data format is shown in Table \ref{fig:format}.
We concatenated the title and chapter title of the literature review, the abstract of the cited paper, and an identifier to distinguish the different cited papers.
They are separated by the token ``<s>,'' and each cited paper's inputs are separated by the token ``</s>.''
In Big Bird, the information on all cited papers is concatenated and input into the model.
In FiD/QFiD, the information on each cited paper is input into the encoder separately.
In LEAD, LexRank, and ext-oracle, only the abstract of each paper is input because titles are noise for the extractive methods.

The models were implemented using PyTorch \cite{NEURIPS2019_bdbca288} and HuggingFace Transformers \cite{wolf-etal-2020-transformers} libraries.
The number of parameters of Big Bird and FiD/QFiD is approximately 577M and 406M, respectively.
These models were trained for ten epochs with a single run, and the final checkpoints were selected based on the ROUGE-2 scores on the validation dataset.
Training required approximately three days on one NVIDIA A100 GPU (40GB).
For validation, 1,000 chapters were randomly sampled from 8,217 chapters in the original validation dataset owing to time constraints.
The AdamW optimizer \cite{Loshchilov2019DecoupledWD} was used as the learning optimizer, with $\beta_1 \!=\! 0.9$, $\beta_2 \!=\! 0.999$, and $\mathrm{learning\_rate} \!=\! 5e-5$.
The model output was decoded by a beam search with $\mathrm{beam\_size}\!=\!4$.
These hyperparameters were determined based on the validation performance.

\section{Experimental Results}

\subsection{Automatic Evaluation}
We report the ROUGE scores \cite{lin-2004-rouge}\footnote{We used the Python implementation of ROUGE (\url{https://github.com/google-research/google-research/tree/master/rouge}) with the option ``use\_stemmer=True''} for the baseline methods and our QFiD on the SciReviewGen dataset in Table \ref{fig:rouge}.
The evaluation results show that FiD-based models outperform the others except for ext-oracle, whereas Big Bird is comparable to LEAD and LexRank.
As Big Bird is pretrained on abstract generation (single document summarization; SDS), it results in significantly lower performance in literature review generation.
These results contrast with those reported in MS\^{}2 and Multi-XScience, where SDS models are competitive with MDS models.
As simply fine-tuning the SDS model does not work, the literature review generation presents distinct characteristics from the above datasets.
In contrast, FiD uses an encoder pretrained on a SDS task and encodes each cited paper separately, leading to significantly higher performance. 
Furthermore, FiD-based models outperform ext-oracle regarding ROUGE-1 and ROUGE-L, which shows the difficulty of our task since simply copying sentences from the the cited papers does not work well.

Table \ref{fig:rouge} show that QFiD outperforms all the baseline methods, including vanilla FiD.
This improvement suggests that QFiD can generate more appropriate reviews by considering the relevance of each cited paper to the queries.

\begin{table}[t]
  \begin{center}
         \scalebox{0.9}{
         \begin{tabular}{lrrr}
          \toprule
          Models & ROUGE-1 & ROUGE-2 & ROUGE-L \\
          \midrule
          LEAD & 23.09 & 4.68 & 11.72 \\  
          LexRank & 24.40 & 5.02 & 12.52 \\  
          Ext-oracle & 29.43 & 10.13 & 14.88 \\
          Big Bird & 24.25 & 4.08 & 15.30 \\
          FiD & 32.40 & 6.75 & 16.17 \\
          QFiD (ours) & \textbf{34.00} & \textbf{7.75} & \textbf{16.52} \\
          \bottomrule
         \end{tabular}
         }
  \end{center}
  \caption{ROUGE evaluation results on SciReviewGen.}
  \label{fig:rouge}
\end{table}

\begin{table*}[t]
  \begin{center}
         \scalebox{0.9}{
         \begin{tabular}{lrrrrr}
          \toprule
          Evaluation results & Relevance & Coherence & Informativeness & Factuality & Overall \\
          \midrule
          Ground truth > Generated & 25.6\% & 48.9\% & 64.4\% & 40.0\% & 68.9\% \\
          Comparable & 56.7\% & 31.1\% & 20.0\% & 48.9\% & 8.9\% \\
          Generated > Ground truth & 17.8\% & 20.0\% & 15.6\% & 11.1\% & 22.2\% \\
          \bottomrule
         \end{tabular}
         }
  \end{center}
  \caption{Human evaluation results on the SciReviewGen dataset. We show the percentage of the ground truth chapters rated superior/comparable/inferior to the chapters generated by QFiD.}
  \label{fig:humaneval}
\end{table*}

\subsection{Human Evaluation\label{humaneval}}
We conducted a human evaluation of QFiD, which yielded the highest ROUGE score.
The generated and ground truth chapters were compared following the five criteria.
\begin{itemize}
    \setlength{\itemsep}{0cm}
    \item \emph{Relevance}: relevance to the title of the paper and chapter
    \item \emph{Coherence}: how well the text is structured and coherent
    \item \emph{Informativeness}: whether the text mentions concrete information in the cited papers, not only general information
    \item \emph{Factuality}: whether the text does not contradict the content of the cited papers
    \item \emph{Overall}: which of the texts is preferable as a literature review?
\end{itemize}

As the evaluation required expert knowledge, we asked three annotators with graduate-level computer science backgrounds to perform the evaluation.
All annotators had at least one year of research experience in computer vision.
We asked them to rate the generated chapters superior, comparable, or inferior to the ground truth chapter according to each criterion.
The generated chapters, ground truth chapters, cited papers' abstracts, cited papers' body text (as needed), literature review titles, and chapter titles were provided for the annotators.
They were not informed which of the two chapters was the ground truth.

We selected five literature reviews in the computer vision domain for the evaluation \cite{wang2020deep, jiao2019survey, hossain2019comprehensive, laga2019survey, tian2020deep}.
All of them had less than 20\% overlap of cited papers with any literature review in the training set.
Since a considerable amount of time is required to evaluate long scientific texts, we randomly selected 30 chapters for the evaluation, where the total number of words in the cited papers' abstracts was less than 1,000, and that of the ground truth was less than 400.
The papers/chapters used for the evaluation were chosen regardless of the quality of the generated text.

Table \ref{fig:humaneval} shows the human evaluation results.
The percentages indicate the proportion of the ground truth chapters that are rated superior to the generated chapters (Ground truth > Generated), comparable, and inferior to the generated chapters (Generated > Ground truth) w.r.t. each criterion.
The inter-annotator agreement is scored as Cohen's kappa = 0.212, which is reasonable because the number of categories is three \cite{hallgren2012computing}.
The ground truth outperforms the generated chapters for all criteria, indicating that automatic literature review generation does not achieve human-level performance.
However, regarding \emph{overall}, 68.9\% of the ground truth outperforms the generated chapters, whereas 22.2\% of the generated chapters outperforms the ground truth.
This result is surprising because some machine-generated chapters are more sophisticated than those written by experts.

The generated chapters achieve relatively high scores for \emph{relevance} and \emph{coherence}.
Specifically, for \emph{relevance}, 74.5\% of the generated chapters are comparable or superior to the ground truth, indicating that our QFiD can generate coherent summaries concerning the titles of papers and chapters.
However, for \emph{informativeness} and \emph{factuality}, the generated chapters remarkably underperform the ground truth.
This underperformance suggests that generated reviews tend to describe general or sometimes incorrect information.
Specifically, while the total percentage of generated chapters comparable or superior to the ground truth is 60.0\% w.r.t. \emph{factuality}, the percentage is only 35.6\% for \emph{informativeness}.
We elaborate on these causes in Section \ref{discussion}. 

Table \ref{fig:ex} shows an example of a chapter describing \emph{progressive upsampling super-resolution}, one of the techniques for upsampling operation.
Two annotators rate the generated chapter superior to the ground truth w.r.t. \emph{overall}.
The ground truth first mentions the general upsampling operation (BIB001 and BIB003) and then explains progressive upsampling super-resolution in detail (BIB002, BIB004, and BIB005). 
In contrast, the generated chapter consistently focuses on progressive upsampling super-resolution by referring to BIB002, BIB004, and BIB005, and explains the details of the papers with sufficient fluency.
This example suggests that the generated chapter appropriately focuses on content related to the titles while maintaining sufficient consistency.
For more examples, see Appendix \ref{sec:appendix1} and \ref{sec:appendix2}.

\begin{table}[t]
\centering
\begin{tabularx}{\columnwidth}{|X|}
\hline
\fontsize{8}{8}\selectfont
Title: \textbf{Deep Learning for Image Super-resolution: A Survey} \\ \hline
\fontsize{8}{8}\selectfont
Chapter title: \textbf{Progressive Upsampling Super-resolution} \\ \hline
\fontsize{8}{8}\selectfont
\textbf{Ground truth chapter} \cite{wang2020deep} \\ \hline
\fontsize{8}{8}\selectfont
Although models under post-upsampling SR framework have reduced the immensely large computational and runtime cost, it still has some shortcomings. On the one hand, the upsampling operation is performed in only one step, which greatly increases the learning difficulty for large scaling factors (e.g., BIB001 BIB003 . On the other hand, each scaling factor requires an individual SR model, which cannot cope with the need for multi-scale SR. To address these drawbacks, a progressive upsampling SR framework is adopted by Laplacian pyramid SR network (LapSRN) BIB002 , as Fig. 2c shows. Specifically, the models under this framework are based on a cascade of CNNs and progressively reconstruct higher-resolution images. At each stage, the images are upsampled to higher resolution and refined by CNNs. Some other works such as MS-LapSRN BIB005 and progressive SR (ProSR) BIB004 also adopt this framework and achieve relatively high performance. ...
\\ \hline
\fontsize{8}{8}\selectfont
\textbf{Generated chapter} \\ \hline
\fontsize{8}{8}\selectfont
The Progressive Upsampling Super-Resolution (PSR) BIB002, BIB004 is proposed to progressively reconstruct the sub-band residuals of high-resolution images at multiple pyramid levels. This method does not require the bicubic interpolation step as the pre-processing step, and thus dramatically reduces the computational complexity. The key idea of PSR is that the network upsamples an image in intermediate steps, while the learning process is organized from easy to hard, as is done in curriculum learning. To obtain more photorealistic results, a generative adversarial network (GAN), named ProGanSR, is proposed, which follows the same progressive multi-scale design principle. This not only allows to scale well to high upsampling factors (e.g., 8×) but also constitutes a principled multiscale approach that increases the reconstruction quality for all up-sampling factors simultaneously BIB005. The authors train the network with deep supervision using the robust Charbonnier loss function and achieve high-quality image reconstruction. Furthermore, the recursive layers are used to share parameters across as well as within pyramid levels, which drastically reduce the number of parameters. In addition, the authors use the recursive loss function to
\\ \hline
\end{tabularx}
\caption{Example of a generated chapter and corresponding ground truth. Two annotators rate the generated chapter superior to the ground truth w.r.t. \emph{overall}.}
\label{fig:ex}
\end{table}

\subsection{Discussions}
\label{discussion}
As shown in Section \ref{humaneval}, the generated chapters considerably underperform the ground truth concerning \emph{informativeness} and \emph{factuality}, which may be attributed to the lack of source information.
Since only abstracts are input into the model, it is difficult to describe the details of the cited papers.
As discussed in \citet{ji2022survey}, hallucinations tend to occur when the target text contains a large amount of information absent from the source. 
Therefore, adding other input information, such as body text, citation sentences, and the text of co-cited papers, will improve both \emph{informativeness} and \emph{factuality}.
In addition to textual information, citation networks can be used to determine which cited papers should be focused on concerning the topic.
While this study uses only abstracts and titles as the first step for literature review generation, using the aforementioned information would be required in future research.

The human evaluation results clarified that we have not yet reached the stage of fully automatic literature review generation without manual modifications.
At the same time, we show some promising results that approximately 30\% of the generated chapters are competitive or superior to human-written reviews concerning \emph{overall}.
This result suggests that a fully automatic generation of literature reviews will be possible if the remaining issues, such as hallucinations and less informativeness, are solved.
Currently, automatic literature review generation can be effectively utilized with human revision, such as writing assistance tools, by providing drafts of literature reviews.

\section{Conclusion}
We propose SciReviewGen, a large-scale dataset for automatic literature review generation.
We also introduce an extension of FiD \cite{izacard-grave-2021-leveraging} for query-focused summarization and show that our QFiD model outperforms naive FiD.
The human evaluation results show that some generated texts are comparable to human-written literature reviews.
However, challenges such as hallucinations and a lack of detailed information still remain to be addressed. 
We hope that our study will serve as a basis for future research on the challenging task of automatic literature review generation.

\section*{Limitations}
In our experiment, we use only abstract text as the input text for literature review generation
However, in writing literature reviews, a writer reads the full text of each cited paper and even other papers related to the research area.
Therefore, the input data are insufficient to write a complete literature review.
As only 70\% of the cited papers have access to the body text in our SciReviewGen, a dataset containing full-text information is required for further research.

In human-written literature reviews, the chapters complement each other and are not redundant.
However, as our QFiD and baseline models generate each chapter independently, they cannot consider the relationships between chapters.
Furthermore, the relations between each cited paper are considered in the actual literature review writing process (e.g., which paper is the first on the topic and which is the following).
However, these relationships are not considered in the models.
In future research, a literature review generation model that can consider the relations between chapters and cited papers by using additional information, such as the contents of other chapters, citation networks, and citation sentences, should be investigated.

As mentioned in Section \ref{humaneval}, the generated text contains incorrect information to a certain extent.
Therefore, we cannot publish it without human revision.
Currently, the model can be utilized as a writing assistance tool, not as a complete literature review generation model.

\section*{Ethics Statement}
\paragraph{Potential Risks}
As discussed in Section \ref{humaneval}, our model risks generating incorrect information.
Currently, it can be effectively utilized with human revision, such as writing assistance tools, by providing drafts of literature reviews.
However, a literature review with wrong information could be published if abused.

\paragraph{Licenses}
We used S2ORC \cite[][CC BY-NC 4.0]{lo-etal-2020-s2orc}, PyTorch \cite[][BSD-style license]{NEURIPS2019_bdbca288}, and HuggingFace Transformers \cite[][MIT for facebook/bart-large-cnn, Apache-2.0 for all materials]{wolf-etal-2020-transformers} as scientific artifacts.
All artifacts can be used for research purposes.
We release the SciReviewGen dataset based on S2ORC, as CC BY-NC 4.0 allows users to adapt and share licensed material for noncommercial purposes.

\paragraph{Annotation Procedures}
The annotation procedures complied with the ACL Ethics Policy.
Prior to the annotation, we informed the ethics review board in our university of the annotation procedures and were notified that it was exempt from ethics review.
More details are presented in Section \ref{appendix3}.

\section*{Acknowledgements}

We would like to thank the anonymous reviewers for their valuable feedback.
This work was supported by NEDO JPNP20006, JST ACT-X JPMJAX1904, JST CREST JPMJCR21D1, Japan.

\bibliography{anthology,custom}

\begin{thebibliography}{44}
\expandafter\ifx\csname natexlab\endcsname\relax\def\natexlab#1{#1}\fi

\bibitem[{Agarwal et~al.(2011)Agarwal, Reddy, Gvr, and
  Ros{\'e}}]{agarwal-etal-2011-towards}
Nitin Agarwal, Ravi~Shankar Reddy, Kiran Gvr, and Carolyn~Penstein Ros{\'e}.
  2011.
\newblock \href {https://aclanthology.org/W11-0502} {Towards multi-document
  summarization of scientific articles:making interesting comparisons with
  {S}ci{S}umm}.
\newblock In \emph{Proceedings of the Workshop on Automatic Summarization for
  Different Genres, Media, and Languages}, pages 8--15, Portland, Oregon.
  Association for Computational Linguistics.

\bibitem[{Beltagy et~al.(2019)Beltagy, Lo, and
  Cohan}]{beltagy-etal-2019-scibert}
Iz~Beltagy, Kyle Lo, and Arman Cohan. 2019.
\newblock \href {https://doi.org/10.18653/v1/D19-1371} {{S}ci{BERT}: A
  pretrained language model for scientific text}.
\newblock In \emph{Proceedings of the 2019 Conference on Empirical Methods in
  Natural Language Processing and the 9th International Joint Conference on
  Natural Language Processing (EMNLP-IJCNLP)}, pages 3615--3620, Hong Kong,
  China. Association for Computational Linguistics.

\bibitem[{Beltagy et~al.(2020)Beltagy, Peters, and
  Cohan}]{beltagy2020longformer}
Iz~Beltagy, Matthew~E Peters, and Arman Cohan. 2020.
\newblock Longformer: The long-document transformer.
\newblock \emph{arXiv preprint arXiv:2004.05150}.

\bibitem[{Bornmann and Mutz(2015)}]{bornmann2015growth}
Lutz Bornmann and R{\"u}diger Mutz. 2015.
\newblock Growth rates of modern science: A bibliometric analysis based on the
  number of publications and cited references.
\newblock \emph{Journal of the Association for Information Science and
  Technology}, 66(11):2215--2222.

\bibitem[{Brown et~al.(2020)Brown, Mann, Ryder, Subbiah, Kaplan, Dhariwal,
  Neelakantan, Shyam, Sastry, Askell, Agarwal, Herbert-Voss, Krueger, Henighan,
  Child, Ramesh, Ziegler, Wu, Winter, Hesse, Chen, Sigler, Litwin, Gray, Chess,
  Clark, Berner, McCandlish, Radford, Sutskever, and
  Amodei}]{brown2020language}
Tom Brown, Benjamin Mann, Nick Ryder, Melanie Subbiah, Jared~D Kaplan, Prafulla
  Dhariwal, Arvind Neelakantan, Pranav Shyam, Girish Sastry, Amanda Askell,
  Sandhini Agarwal, Ariel Herbert-Voss, Gretchen Krueger, Tom Henighan, Rewon
  Child, Aditya Ramesh, Daniel Ziegler, Jeffrey Wu, Clemens Winter, Chris
  Hesse, Mark Chen, Eric Sigler, Mateusz Litwin, Scott Gray, Benjamin Chess,
  Jack Clark, Christopher Berner, Sam McCandlish, Alec Radford, Ilya Sutskever,
  and Dario Amodei. 2020.
\newblock \href
  {https://proceedings.neurips.cc/paper/2020/file/1457c0d6bfcb4967418bfb8ac142f64a-Paper.pdf}
  {Language models are few-shot learners}.
\newblock In \emph{Advances in Neural Information Processing Systems},
  volume~33, pages 1877--1901.

\bibitem[{Cohan et~al.(2018)Cohan, Dernoncourt, Kim, Bui, Kim, Chang, and
  Goharian}]{cohan-etal-2018-discourse}
Arman Cohan, Franck Dernoncourt, Doo~Soon Kim, Trung Bui, Seokhwan Kim, Walter
  Chang, and Nazli Goharian. 2018.
\newblock \href {https://doi.org/10.18653/v1/N18-2097} {A discourse-aware
  attention model for abstractive summarization of long documents}.
\newblock In \emph{Proceedings of the 2018 Conference of the North {A}merican
  Chapter of the Association for Computational Linguistics: Human Language
  Technologies, Volume 2 (Short Papers)}, pages 615--621, New Orleans,
  Louisiana. Association for Computational Linguistics.

\bibitem[{Cohan et~al.(2022)Cohan, Feigenblat, Freitag, Ghosal, Herrmannova,
  Knoth, Lo, Mayr, Shmueli-Scheuer, de~Waard, and
  Wang}]{cohan-etal-2022-overview}
Arman Cohan, Guy Feigenblat, Dayne Freitag, Tirthankar Ghosal, Drahomira
  Herrmannova, Petr Knoth, Kyle Lo, Philipp Mayr, Michal Shmueli-Scheuer, Anita
  de~Waard, and Lucy~Lu Wang. 2022.
\newblock \href {https://aclanthology.org/2022.sdp-1.1} {Overview of the third
  workshop on scholarly document processing}.
\newblock In \emph{Proceedings of the Third Workshop on Scholarly Document
  Processing}, pages 1--6, Gyeongju, Republic of Korea. Association for
  Computational Linguistics.

\bibitem[{Cohan et~al.(2020)Cohan, Feldman, Beltagy, Downey, and
  Weld}]{cohan-etal-2020-specter}
Arman Cohan, Sergey Feldman, Iz~Beltagy, Doug Downey, and Daniel Weld. 2020.
\newblock \href {https://doi.org/10.18653/v1/2020.acl-main.207} {{SPECTER}:
  Document-level representation learning using citation-informed transformers}.
\newblock In \emph{Proceedings of the 58th Annual Meeting of the Association
  for Computational Linguistics}, pages 2270--2282, Online. Association for
  Computational Linguistics.

\bibitem[{DeYoung et~al.(2021)DeYoung, Beltagy, van Zuylen, Kuehl, and
  Wang}]{deyoung-etal-2021-ms}
Jay DeYoung, Iz~Beltagy, Madeleine van Zuylen, Bailey Kuehl, and Lucy~Lu Wang.
  2021.
\newblock \href {https://doi.org/10.18653/v1/2021.emnlp-main.594} {{MS}{\^{}}2:
  Multi-document summarization of medical studies}.
\newblock In \emph{Proceedings of the 2021 Conference on Empirical Methods in
  Natural Language Processing}, pages 7494--7513, Online and Punta Cana,
  Dominican Republic. Association for Computational Linguistics.

\bibitem[{Erkan and Radev(2004)}]{erkan2004lexrank}
G{\"u}nes Erkan and Dragomir~R Radev. 2004.
\newblock Lexrank: Graph-based lexical centrality as salience in text
  summarization.
\newblock \emph{Journal of artificial intelligence research}, 22:457--479.

\bibitem[{Fabbri et~al.(2019)Fabbri, Li, She, Li, and
  Radev}]{fabbri-etal-2019-multi}
Alexander Fabbri, Irene Li, Tianwei She, Suyi Li, and Dragomir Radev. 2019.
\newblock \href {https://doi.org/10.18653/v1/P19-1102} {Multi-news: A
  large-scale multi-document summarization dataset and abstractive hierarchical
  model}.
\newblock In \emph{Proceedings of the 57th Annual Meeting of the Association
  for Computational Linguistics}, pages 1074--1084, Florence, Italy.
  Association for Computational Linguistics.

\bibitem[{Hallgren(2012)}]{hallgren2012computing}
Kevin~A Hallgren. 2012.
\newblock Computing inter-rater reliability for observational data: an overview
  and tutorial.
\newblock \emph{Tutorials in quantitative methods for psychology}, 8(1):23--34.

\bibitem[{Hossain et~al.(2019)Hossain, Sohel, Shiratuddin, and
  Laga}]{hossain2019comprehensive}
MD~Zakir Hossain, Ferdous Sohel, Mohd~Fairuz Shiratuddin, and Hamid Laga. 2019.
\newblock A comprehensive survey of deep learning for image captioning.
\newblock \emph{ACM Computing Surveys}, 51(6):1--36.

\bibitem[{Izacard and Grave(2021)}]{izacard-grave-2021-leveraging}
Gautier Izacard and Edouard Grave. 2021.
\newblock \href {https://doi.org/10.18653/v1/2021.eacl-main.74} {Leveraging
  passage retrieval with generative models for open domain question answering}.
\newblock In \emph{Proceedings of the 16th Conference of the European Chapter
  of the Association for Computational Linguistics: Main Volume}, pages
  874--880, Online. Association for Computational Linguistics.

\bibitem[{Jaidka et~al.(2013{\natexlab{a}})Jaidka, Khoo, and
  Na}]{jaidka-etal-2013-deconstructing}
Kokil Jaidka, Christopher Khoo, and Jin-Cheon Na. 2013{\natexlab{a}}.
\newblock \href {https://aclanthology.org/W13-2116} {Deconstructing human
  literature reviews {--} a framework for multi-document summarization}.
\newblock In \emph{Proceedings of the 14th {E}uropean Workshop on Natural
  Language Generation}, pages 125--135, Sofia, Bulgaria. Association for
  Computational Linguistics.

\bibitem[{Jaidka et~al.(2013{\natexlab{b}})Jaidka, Khoo, and
  Na}]{jaidka2013literature}
Kokil Jaidka, Christopher~SG Khoo, and Jin-Cheon Na. 2013{\natexlab{b}}.
\newblock Literature review writing: how information is selected and
  transformed.
\newblock In \emph{Aslib Proceedings}, volume~65, pages 303--325.

\bibitem[{Jaidka et~al.(2019)Jaidka, Khoo, and Na}]{jaidka2019characterizing}
Kokil Jaidka, Christopher~SG Khoo, and Jin-Cheon Na. 2019.
\newblock Characterizing human summarization strategies for text reuse and
  transformation in literature review writing.
\newblock \emph{Scientometrics}, 121(3):1563--1582.

\bibitem[{Ji et~al.(2022)Ji, Lee, Frieske, Yu, Su, Xu, Ishii, Bang, Madotto,
  and Fung}]{ji2022survey}
Ziwei Ji, Nayeon Lee, Rita Frieske, Tiezheng Yu, Dan Su, Yan Xu, Etsuko Ishii,
  Yejin Bang, Andrea Madotto, and Pascale Fung. 2022.
\newblock Survey of hallucination in natural language generation.
\newblock \emph{ACM Computing Surveys}.
\newblock Just Accepted.

\bibitem[{Jiao and Zhao(2019)}]{jiao2019survey}
Licheng Jiao and Jin Zhao. 2019.
\newblock A survey on the new generation of deep learning in image processing.
\newblock \emph{IEEE Access}, 7:172231--172263.

\bibitem[{Khan et~al.(2003)Khan, Kunz, Kleijnen, and Antes}]{khan2003five}
Khalid~S Khan, Regina Kunz, Jos Kleijnen, and Gerd Antes. 2003.
\newblock Five steps to conducting a systematic review.
\newblock \emph{Journal of the royal society of medicine}, 96(3):118--121.

\bibitem[{Laga(2019)}]{laga2019survey}
Hamid Laga. 2019.
\newblock A survey on deep learning architectures for image-based depth
  reconstruction.
\newblock \emph{arXiv preprint arXiv:1906.06113}.

\bibitem[{Laskar et~al.(2022)Laskar, Hoque, and
  Huang}]{laskar-etal-2022-domain}
Md~Tahmid~Rahman Laskar, Enamul Hoque, and Jimmy~Xiangji Huang. 2022.
\newblock \href {https://doi.org/10.1162/coli_a_00434} {Domain adaptation with
  pre-trained transformers for query-focused abstractive text summarization}.
\newblock \emph{Computational Linguistics}, 48(2):279--320.

\bibitem[{LeCun et~al.(1989)LeCun, Boser, Denker, Henderson, Howard, Hubbard,
  and Jackel}]{lecun1989backpropagation}
Yann LeCun, Bernhard Boser, John~S Denker, Donnie Henderson, Richard~E Howard,
  Wayne Hubbard, and Lawrence~D Jackel. 1989.
\newblock Backpropagation applied to handwritten zip code recognition.
\newblock \emph{Neural computation}, 1(4):541--551.

\bibitem[{Lewis et~al.(2020)Lewis, Liu, Goyal, Ghazvininejad, Mohamed, Levy,
  Stoyanov, and Zettlemoyer}]{lewis-etal-2020-bart}
Mike Lewis, Yinhan Liu, Naman Goyal, Marjan Ghazvininejad, Abdelrahman Mohamed,
  Omer Levy, Veselin Stoyanov, and Luke Zettlemoyer. 2020.
\newblock \href {https://doi.org/10.18653/v1/2020.acl-main.703} {{BART}:
  Denoising sequence-to-sequence pre-training for natural language generation,
  translation, and comprehension}.
\newblock In \emph{Proceedings of the 58th Annual Meeting of the Association
  for Computational Linguistics}, pages 7871--7880, Online. Association for
  Computational Linguistics.

\bibitem[{Lin(2004)}]{lin-2004-rouge}
Chin-Yew Lin. 2004.
\newblock \href {https://aclanthology.org/W04-1013} {{ROUGE}: A package for
  automatic evaluation of summaries}.
\newblock In \emph{Text Summarization Branches Out}, pages 74--81, Barcelona,
  Spain. Association for Computational Linguistics.

\bibitem[{Liu et~al.(2019)Liu, Shi, Ji, and Jia}]{liu2019survey}
Ruijun Liu, Yuqian Shi, Changjiang Ji, and Ming Jia. 2019.
\newblock \href {https://doi.org/10.1109/ACCESS.2019.2925059} {A survey of
  sentiment analysis based on transfer learning}.
\newblock \emph{IEEE Access}, 7:85401--85412.

\bibitem[{Lo et~al.(2020)Lo, Wang, Neumann, Kinney, and
  Weld}]{lo-etal-2020-s2orc}
Kyle Lo, Lucy~Lu Wang, Mark Neumann, Rodney Kinney, and Daniel Weld. 2020.
\newblock \href {https://doi.org/10.18653/v1/2020.acl-main.447} {{S}2{ORC}: The
  semantic scholar open research corpus}.
\newblock In \emph{Proceedings of the 58th Annual Meeting of the Association
  for Computational Linguistics}, pages 4969--4983, Online. Association for
  Computational Linguistics.

\bibitem[{Loshchilov and Hutter(2019)}]{Loshchilov2019DecoupledWD}
Ilya Loshchilov and Frank Hutter. 2019.
\newblock Decoupled weight decay regularization.
\newblock In \emph{International Conference on Learning Representations}.

\bibitem[{Lu et~al.(2020)Lu, Dong, and Charlin}]{lu-etal-2020-multi-xscience}
Yao Lu, Yue Dong, and Laurent Charlin. 2020.
\newblock \href {https://doi.org/10.18653/v1/2020.emnlp-main.648}
  {Multi-{XS}cience: A large-scale dataset for extreme multi-document
  summarization of scientific articles}.
\newblock In \emph{Proceedings of the 2020 Conference on Empirical Methods in
  Natural Language Processing (EMNLP)}, pages 8068--8074, Online. Association
  for Computational Linguistics.

\bibitem[{Mohammad et~al.(2009)Mohammad, Dorr, Egan, Hassan, Muthukrishan,
  Qazvinian, Radev, and Zajic}]{mohammad-etal-2009-using}
Saif Mohammad, Bonnie Dorr, Melissa Egan, Ahmed Hassan, Pradeep Muthukrishan,
  Vahed Qazvinian, Dragomir Radev, and David Zajic. 2009.
\newblock \href {https://aclanthology.org/N09-1066} {Using citations to
  generate surveys of scientific paradigms}.
\newblock In \emph{Proceedings of Human Language Technologies: The 2009 Annual
  Conference of the North {A}merican Chapter of the Association for
  Computational Linguistics}, pages 584--592, Boulder, Colorado. Association
  for Computational Linguistics.

\bibitem[{Nallapati et~al.(2016)Nallapati, Zhou, dos Santos, Gul{\c{c}}ehre,
  and Xiang}]{nallapati-etal-2016-abstractive}
Ramesh Nallapati, Bowen Zhou, Cicero dos Santos, {\c{C}}a{\u{g}}lar
  Gul{\c{c}}ehre, and Bing Xiang. 2016.
\newblock \href {https://doi.org/10.18653/v1/K16-1028} {Abstractive text
  summarization using sequence-to-sequence {RNN}s and beyond}.
\newblock In \emph{Proceedings of the 20th {SIGNLL} Conference on Computational
  Natural Language Learning}, pages 280--290, Berlin, Germany. Association for
  Computational Linguistics.

\bibitem[{Narayan et~al.(2018)Narayan, Cohen, and
  Lapata}]{narayan-etal-2018-dont}
Shashi Narayan, Shay~B. Cohen, and Mirella Lapata. 2018.
\newblock \href {https://doi.org/10.18653/v1/D18-1206} {Don{'}t give me the
  details, just the summary! topic-aware convolutional neural networks for
  extreme summarization}.
\newblock In \emph{Proceedings of the 2018 Conference on Empirical Methods in
  Natural Language Processing}, pages 1797--1807, Brussels, Belgium.
  Association for Computational Linguistics.

\bibitem[{PAGE(1998)}]{page1998pagerank}
L~PAGE. 1998.
\newblock The pagerank citation ranking: Bringing order to the web.
\newblock In \emph{Proc. of the 7\^{}< th> WWW Conf., 1998}.

\bibitem[{Paszke et~al.(2019)Paszke, Gross, Massa, Lerer, Bradbury, Chanan,
  Killeen, Lin, Gimelshein, Antiga, Desmaison, Kopf, Yang, DeVito, Raison,
  Tejani, Chilamkurthy, Steiner, Fang, Bai, and
  Chintala}]{NEURIPS2019_bdbca288}
Adam Paszke, Sam Gross, Francisco Massa, Adam Lerer, James Bradbury, Gregory
  Chanan, Trevor Killeen, Zeming Lin, Natalia Gimelshein, Luca Antiga, Alban
  Desmaison, Andreas Kopf, Edward Yang, Zachary DeVito, Martin Raison, Alykhan
  Tejani, Sasank Chilamkurthy, Benoit Steiner, Lu~Fang, Junjie Bai, and Soumith
  Chintala. 2019.
\newblock \href
  {https://proceedings.neurips.cc/paper/2019/file/bdbca288fee7f92f2bfa9f7012727740-Paper.pdf}
  {Pytorch: An imperative style, high-performance deep learning library}.
\newblock In \emph{Advances in Neural Information Processing Systems},
  volume~32, pages 8026--8037.

\bibitem[{Taylor et~al.(2022{\natexlab{a}})Taylor, Kardas, Cucurull, Scialom,
  Hartshorn, Saravia, Poulton, Kerkez, and Stojnic}]{taylor2022galactica}
Ross Taylor, Marcin Kardas, Guillem Cucurull, Thomas Scialom, Anthony
  Hartshorn, Elvis Saravia, Andrew Poulton, Viktor Kerkez, and Robert Stojnic.
  2022{\natexlab{a}}.
\newblock Galactica: A large language model for science.
\newblock \emph{arXiv preprint arXiv:2211.09085}.

\bibitem[{Taylor et~al.(2022{\natexlab{b}})Taylor, Kardas, Cucurull, Scialom,
  Hartshorn, Saravia, Poulton, Kerkez, and Stojnic}]{Galactic18:online}
Ross Taylor, Marcin Kardas, Guillem Cucurull, Thomas Scialom, Anthony
  Hartshorn, Elvis Saravia, Andrew Poulton, Viktor Kerkez, and Robert Stojnic.
  2022{\natexlab{b}}.
\newblock Galactica demo.
\newblock \url{https://galactica.org/}.
\newblock (Accessed on 12/26/2022).

\bibitem[{Tian et~al.(2020)Tian, Fei, Zheng, Xu, Zuo, and Lin}]{tian2020deep}
Chunwei Tian, Lunke Fei, Wenxian Zheng, Yong Xu, Wangmeng Zuo, and Chia-Wen
  Lin. 2020.
\newblock Deep learning on image denoising: An overview.
\newblock \emph{Neural Networks}, 131:251--275.

\bibitem[{Vaswani et~al.(2017)Vaswani, Shazeer, Parmar, Uszkoreit, Jones,
  Gomez, Kaiser, and Polosukhin}]{vaswani2017attention}
Ashish Vaswani, Noam Shazeer, Niki Parmar, Jakob Uszkoreit, Llion Jones,
  Aidan~N Gomez, \L~ukasz Kaiser, and Illia Polosukhin. 2017.
\newblock \href
  {https://proceedings.neurips.cc/paper/2017/file/3f5ee243547dee91fbd053c1c4a845aa-Paper.pdf}
  {Attention is all you need}.
\newblock In \emph{Advances in Neural Information Processing Systems},
  volume~30, pages 6000--–6010.

\bibitem[{Vig et~al.(2022)Vig, Fabbri, Kryscinski, Wu, and
  Liu}]{vig-etal-2022-exploring}
Jesse Vig, Alexander Fabbri, Wojciech Kryscinski, Chien-Sheng Wu, and Wenhao
  Liu. 2022.
\newblock \href {https://doi.org/10.18653/v1/2022.findings-naacl.109}
  {Exploring neural models for query-focused summarization}.
\newblock In \emph{Findings of the Association for Computational Linguistics:
  NAACL 2022}, pages 1455--1468, Seattle, United States. Association for
  Computational Linguistics.

\bibitem[{Wang et~al.(2020)Wang, Chen, and Hoi}]{wang2020deep}
Zhihao Wang, Jian Chen, and Steven~CH Hoi. 2020.
\newblock Deep learning for image super-resolution: A survey.
\newblock \emph{IEEE transactions on pattern analysis and machine
  intelligence}, 43(10):3365--3387.

\bibitem[{Wolf et~al.(2020)Wolf, Debut, Sanh, Chaumond, Delangue, Moi, Cistac,
  Rault, Louf, Funtowicz, Davison, Shleifer, von Platen, Ma, Jernite, Plu, Xu,
  Le~Scao, Gugger, Drame, Lhoest, and Rush}]{wolf-etal-2020-transformers}
Thomas Wolf, Lysandre Debut, Victor Sanh, Julien Chaumond, Clement Delangue,
  Anthony Moi, Pierric Cistac, Tim Rault, Remi Louf, Morgan Funtowicz, Joe
  Davison, Sam Shleifer, Patrick von Platen, Clara Ma, Yacine Jernite, Julien
  Plu, Canwen Xu, Teven Le~Scao, Sylvain Gugger, Mariama Drame, Quentin Lhoest,
  and Alexander Rush. 2020.
\newblock \href {https://doi.org/10.18653/v1/2020.emnlp-demos.6} {Transformers:
  State-of-the-art natural language processing}.
\newblock In \emph{Proceedings of the 2020 Conference on Empirical Methods in
  Natural Language Processing: System Demonstrations}, pages 38--45, Online.
  Association for Computational Linguistics.

\bibitem[{Yasunaga et~al.(2019)Yasunaga, Kasai, Zhang, Fabbri, Li, Friedman,
  and Radev}]{yasunaga2019scisummnet}
Michihiro Yasunaga, Jungo Kasai, Rui Zhang, Alexander~R Fabbri, Irene Li, Dan
  Friedman, and Dragomir~R Radev. 2019.
\newblock Scisummnet: A large annotated corpus and content-impact models for
  scientific paper summarization with citation networks.
\newblock In \emph{Proceedings of the AAAI conference on artificial
  intelligence}, volume~33, pages 7386--7393.

\bibitem[{Yuan et~al.(2022)Yuan, Liu, and Neubig}]{yuan2022can}
Weizhe Yuan, Pengfei Liu, and Graham Neubig. 2022.
\newblock Can we automate scientific reviewing?
\newblock \emph{Journal of Artificial Intelligence Research}, 75:171--212.

\bibitem[{Zaheer et~al.(2020)Zaheer, Guruganesh, Dubey, Ainslie, Alberti,
  Ontanon, Pham, Ravula, Wang, Yang, and Ahmed}]{zaheer2020big}
Manzil Zaheer, Guru Guruganesh, Kumar~Avinava Dubey, Joshua Ainslie, Chris
  Alberti, Santiago Ontanon, Philip Pham, Anirudh Ravula, Qifan Wang, Li~Yang,
  and Amr Ahmed. 2020.
\newblock \href
  {https://proceedings.neurips.cc/paper/2020/file/c8512d142a2d849725f31a9a7a361ab9-Paper.pdf}
  {Big bird: Transformers for longer sequences}.
\newblock In \emph{Advances in Neural Information Processing Systems},
  volume~33, pages 17283--17297.

\end{thebibliography}
\bibliographystyle{acl_natbib}

\clearpage
\appendix

\section{Examples of generated chapters}
\label{sec:appendix1}
Table \ref{fig:o-relcoh} shows an example of a chapter describing \emph{stylized caption}, one of the categories of image captioning methods.
The generated chapter is rated competitive or superior to the ground truth w.r.t. \emph{relevance} and \emph{coherence} by all annotators.
The first half describes the background of stylized captions, while the second half describes the details of BIB002 and BIB003, which are both methods of stylized captions.
This example suggests that the generated chapter is a consistent and structured summary of stylized captions.

Table \ref{fig:x-inffac} shows an example of a chapter describing \emph{Convolutional neural networks (CNN)}.
The generated chapter is rated inferior to the ground truth w.r.t. \emph{informativeness} and \emph{factuality} by more than two annotators.
The generated chapter does not refer to BIB002 and BIB003 and contains only general descriptions of CNN.
Moreover, it has a wrong description that BIB001 proposed CNN in 2012.
In fact, CNN was first proposed by \citet{lecun1989backpropagation}, and BIB001 proposed new pooling methods for CNN in 2014.
On the other hand, it correctly states that Yann LeCun proposed CNN, although no input documents state it at all.
This result indicates that the model learned knowledge about the computer vision domain during the training process and includes it in the generated text correctly.

\section{Example of a generated literature review}
\label{sec:appendix2}
We show an example of a literature review generated based on \citet{liu2019survey} at the end of the appendix.
It has less than 20\% overlap of cited papers with any literature review in the training set.
Only chapters that have access to two or more cited papers are shown.

\section{Details of Annotation Procedures\label{appendix3}}
\paragraph{Details of Annotation for Filtering Literature Reviews}
The full instructions to the participants are shown in Table \ref{fig:inst1}.
We recruited three graduate students from our graduate school with graduate-level computer science backgrounds as annotators.
The working hours averaged 50 hours for each annotator, and we paid 100,000 yen as rewards.
The hourly wage is determined according to the university's rules and is higher than the minimum wage in our country.
We informed the annotators that the data would be used to create the SciReviewGen dataset.

\paragraph{Details of Human Evaluation}
The full instructions to the participants are shown in Table \ref{fig:inst2}.
We recruited three graduate students from our graduate school with graduate-level computer science backgrounds for the evaluation.
The working hours averaged 30 hours for each annotator, and we paid 50,000 yen as rewards.
The hourly wage is determined according to the university's rules and is higher than the minimum wage in our country.
We informed the annotators that the data would be used to evaluate the performance of our model, and the evaluation results would be reported herein.

\begin{table}[t]
\centering
\begin{tabularx}{\columnwidth}{|X|}
\hline
\fontsize{9}{9}\selectfont
Referring to the titles and abstracts of 889 candidate papers, please annotate them per the criteria below. Please feel free to ask me if you have any questions.
\begin{itemize}
    \setlength{\leftskip}{-10pt} 
    \setlength{\parskip}{0pt}
    \item Reviewing multiple scientific papers.
    \begin{itemize}
    \setlength{\leftskip}{-20pt}
        \item Not reviewing general tools or books.
        \item Not explaining a specific project or shared task.
    \end{itemize}
    \item Only reviewing scientific papers. Not proposing new methods, re-testing previous studies, or conducting questionnaires (i.e., the paper does not contain content that cannot be generated only by the cited papers' information).
\vspace{-1\baselineskip}
\end{itemize}
\\ \hline
\end{tabularx}
\caption{Full instructions to participants in the suitability annotation}
\label{fig:inst1}

\centering
\begin{tabularx}{\columnwidth}{|X|}
\hline
\fontsize{9}{9}\selectfont
Please evaluate the chapters of literature reviews automatically generated by the model.
A literature review is a scientific paper that summarizes existing scientific articles and provides an overview of the research field. We developed a model that takes the abstracts of the papers cited by the chapter and the titles of the paper and chapter as inputs and generates the chapter. 

Specifically, please evaluate which is better or comparable regarding the generated and human-written chapters following the five criteria below.
We provide the abstracts and full text of the cited papers, the titles of the papers and chapters, the generated chapters, and the human-written chapters.
\begin{itemize}
\setlength{\leftskip}{-15pt}
\setlength{\parskip}{0pt} 
    \item Relevance: Relevance to the title of the paper and chapter.
    \item Coherence: How well the text is structured and coherent.
    \item Informativeness: Whether the text mentions concrete information in the cited papers, not only general information.
    \item Factuality: Whether the text does not contradict the content of the cited papers.
    \item Overall: Which of the texts is preferable as a literature review?
\vspace{-1\baselineskip}
\end{itemize} 
\\ \hline
\end{tabularx}
\caption{Full instructions to participants in the human evaluation}
\label{fig:inst2}
\end{table}

\clearpage

\begin{table}[H]
\centering
\begin{tabularx}{\columnwidth}{|X|}
\hline
\fontsize{8}{8}\selectfont
Title: \textbf{A Comprehensive Survey of Deep Learning for Image Captioning} \\ \hline
\fontsize{8}{8}\selectfont
Chapter title: \textbf{Stylized Caption.} \\ \hline
\fontsize{8}{8}\selectfont
\textbf{Ground truth chapter} \cite{hossain2019comprehensive} \\ \hline
\fontsize{8}{8}\selectfont
Existing image captioning systems generate captions based on only the image content, which can also be called factual descriptions. They do not consider the stylized part of the text separately from other linguistic patterns. However, the stylized captions can be more expressive and attractive than only the flat description of an image. ... Gan et al. BIB003 proposed a novel image captioning system called StyleNet. This method can generate attractive captions, adding various styles. The architecture of this method consists of a CNN and a factored LSTM that can separate factual and style factors from the captions. It uses multitask sequence-to-sequence training BIB001 for identifying the style factors and then adds these factors at runtime to generate attractive captions. ... In our day-to-day conversations, communications, interpersonal relationships, and decision making, we use various stylized and nonfactual expressions such as emotions, pride, and shame. However, Mathews et al. BIB002 claimed that automatic image descriptions are missing these nonfactual aspects. Therefore, they proposed a method called SentiCap. This method can generate image descriptions with positive or negative sentiments. It introduces a novel switching RNN model that combines two CNN+RNNs running in parallel. In each time step, this switching model generates the probability of switching between two RNNs. One generates captions considering the factual words and the other considers the words with sentiments. It then takes inputs from the hidden states of both RNNs for generating captions. This method can generate captions successfully given the appropriate sentiments.
\\ \hline
\fontsize{8}{8}\selectfont
\textbf{Generated chapter} \\ \hline
\fontsize{8}{8}\selectfont
The stylized aspect of image description is missing in current deep learning based image captioning systems. The reason is that these systems focus on the non-factual aspects of the written description, which are missing from the current systems. One such style is descriptions with emotions, which is common in everyday communication, and influences decision-making and interpersonal relationships. Chen et al. BIB002 proposed a model that automatically generates captions with positive or negative sentiments. They use a novel switching recurrent neural network with word-level regularization to produce emotional image captions using only 2000+ training sentences containing sentiments. Their model is evaluated with different automatic and crowd-sourcing metrics. In 84.6\% of cases, the generated positive captions were judged as being at least as descriptive as the factual captions. Another work BIB003 proposed a novel framework named StyleNet to address the task of generating attractive visual captions for images and videos with different styles. The framework consists of two sets of data: factual image/video-caption paired data and stylized monolingual text data (e.g., romantic and humorous sentences). They show experimentally that StyleNet outperforms existing approaches for generating visual captioning with different visual styles, measured in both automatic
\\ \hline
\end{tabularx}
\caption{Example of a generated chapter and corresponding ground truth. All annotators rated the generated chapter comparable or superior to the ground truth w.r.t. \emph{relevance} and \emph{coherence}.}
\label{fig:o-relcoh}
\end{table}

\begin{table}[H]
\centering
\begin{tabularx}{\columnwidth}{|X|}
\hline
\fontsize{8}{8}\selectfont
Title: \textbf{A Survey on the New Generation of Deep Learning in Image Processing} \\ \hline
\fontsize{8}{8}\selectfont
Chapter title: \textbf{A. CNN SERIES MODELS FOR IMAGE PROCESSING 1) CNN} \\ \hline
\fontsize{8}{8}\selectfont
\textbf{Ground truth chapter} \cite{jiao2019survey} \\ \hline
\fontsize{8}{8}\selectfont
In recent years, CNN has also made great success in image processing and object recognition. The strength of CNN lies in their shared weights. Weight sharing dramatically reduces the number of free parameters learned, thus to lower the memory requirements for running the network and allowing the training of more extensive, more powerful networks BIB002. A CNN consists of convolutional layers, pooling layers, normalization layers, and fully connected layers. At each layer, the input image $\boldsymbol{X} \in \mathbb{R}^{n \times m}$ is convolved with a set of $K$ kernels $\left\{\boldsymbol{W}_k \in \mathbb{R}^{v \times v}, k=1,2, \cdots, K\right\}$ and subsequently biases $\left\{b_k \in \mathbb{R}, k=1,2, \cdots, K\right\}$ are added, each generating a new feature map $\boldsymbol{X}_k$ by an element-wise non-linear transform $\sigma(\cdot)$. The same process is repeated for convolutional layer $l$,
$
\boldsymbol{X}_k^l=\sigma\left(\boldsymbol{W}_k^l \otimes \boldsymbol{X}^{l-1}+b_k^l\right)
$
where symbol ' $\otimes$ ' denotes the discrete convolution operator, and its specific type of operation has a variety of forms, such as 'valid' convolution, 'same' convolution, 'extra' convolution, strided convolution, fractional-strided convolution, etc. Another essential layer of CNN is pooling, which is a form of non-linear down-sampling. Convolutional layers are typically alternated with pooling layers where pixel values of neighborhoods are aggregated using some permutation invariant function, usually the max or average operations, which provides another form of translation invariance BIB003-BIB001.
$
\boldsymbol{S}_k^{(l)}=\operatorname{Pooling}\left(\boldsymbol{X}_k^{(l)}\right)
$
Finally, after several convolutional and max-pooling layers, the high-level reasoning in the neural network is done via fully connected layers, where weights are no longer shared. CNN is typically trained end-to-end in an entirely supervised manner. The significant reduction in the number of weights parameters and the translational invariance of the learned features contributes to the ability of $\mathrm{CNN}$ to be trained end-to-end.
\\ \hline
\fontsize{8}{8}\selectfont
\textbf{Generated chapter} \\ \hline
\fontsize{8}{8}\selectfont
The convolutional neural network (CNN) is a special type of feed-forward neural network proposed by Yann LeCun et al. BIB001 in 2012. CNN is a feed forward neural network that consists of a stack of convolution layers, pooling layers, and non-linearity layers. The convolution and pooling operations are performed at the same time in each layer of the CNN. The basic idea of the convolution operation is to convolve the input image with a set of filters at the beginning of the network. Then, the filters are applied to the image in order to extract the feature maps. The feature maps are then passed to the next convolution layer for further processing. Finally, the output is fed to the fully connected layer for classification. The CNN architecture is shown in Fig. 1. Fig. 2 shows the basic architecture of CNN. Fig. 3 shows the general block diagram of CNN, where convolution, max-pooling, average pooling, and fully connected layers are the basic building blocks of the architecture. The architecture of a typical CNN consists of three parts: input, hidden, and output layer. The input layer is used to extract features from the image, and the hidden layers are used
\\ \hline
\end{tabularx}
\caption{Example of a generated chapter and corresponding ground truth. All annotators rated the generated chapter comparable or inferior to the ground truth w.r.t. \emph{informativeness} and \emph{factuality}.}
\label{fig:x-inffac}
\end{table}

\begin{table*}[t]
  \begin{center}
    \begin{tabular}{c}
    \Large
    \textbf{A Survey of Sentiment Analysis Based on Transfer Learning}
    \normalsize \\ \\ \\ \textbf{Generated based on \citet{liu2019survey}}
    \vspace{10pt}
    \end{tabular}
  \end{center}
\end{table*}

\newpage


\label{fullex1}

\section*{I. INTRODUCTION\footnote{As discussed in Section \ref{humaneval}, the generated text contains incorrect information to a certain extent.} }
Sentiment Analysis (SA) is the process of analyzing people's opinions, attitudes, emotions, and emotions expressed in social media data, such as reviews, blogs, forum discussions, and news articles BIB01 - BIB02. Sentiment analysis is an important part of Information Retrieval (IR) and Natural Language Processing (NLP) BIB03. However, it is difficult to obtain a large amount of human-labeled data for Sentiment Analysis because of the huge amount of user-generated content on social media. Therefore, transfer learning has been proposed to overcome this problem and improve the performance of sentiment analysis. Transfer learning is a learning method that transfers knowledge from a source domain to a target domain BIB04. In other words, the target domain is different from the source domain but similar enough in some aspects. For example, if we have a dataset of reviews about a product, we can transfer the reviews about this product to the review about another product. On the other hand, if the dataset of a news event is different but similar in many aspects, then transfer learning can be used to solve the review classification problem. In this paper, we mainly focus on the transfer learning methods for sentiment analysis based on machine learning. Transfer

\section*{II. THE RELATED METHODS}
In this section, we will introduce several related methods that are used to solve the problems in the field of sentiment analysis, including opinion mining BIB05, transfer learning and multi-task learning. These methods can be divided into two main categories: unsupervised learning methods and supervised learning methods. In this paper, we mainly introduce these methods because they have been widely used in sentiment analysis research. According to the taxonomy in BIB06, these methods are divided into three categories: dictionary-based, rule-based and feature-based methods.

\subsection*{A. TRADITIONAL SENTIMENT ANALYSIS}
Sentiment analysis is the process of analyzing opinions and sentiments expressed in natural language text. Sentiment analysis can be divided into two main categories, which are sentiment polarity and sentiment orientation. The former is the evaluation of positive or negative sentiments, and the latter is the interpretation of the sentiment of a word. In general, the polarity of a sentiment word can be positive, negative, or neutral. For example, the word "amazing" can be interpreted as expressingconstructive or negative opinion, and "good" and "bad" are interpreted asconstructive and negative opinions, respectively. On the other hand, the sentiment orientation can be used to indicate negative or positive sentiments. The polarity can be either positive (e.g., great, excellent, excellent) or negative BIB07, BIB08. The sentiment orientation of a sentence can be expressed by a single word or a set of words. For instance, the sentence "I like this camera, but it is not free of bugs" is negative because it expresses strong negative sentiment. The sentiment orientations of the words in a sentence are usually determined by the word co-occurrences in the sentence. Therefore, it is necessary to determine the semantic orientation of each word in

\subsection*{B. SENTIMENT ANALYSIS BASED ON DEEP LEARNING}
In recent years, deep learning has achieved great success in many fields such as computer vision, natural language processing (NLP), speech recognition, and computer vision. The deep learning-based sentiment analysis based on deep neural networks (DNN) BIB09 - BIB10 has been proposed to solve the problems of sentiment classification and sentiment analysis. In this section, we will introduce the deep transfer learning based sentiment analysis methods in the field of NLP and deep learning. Deep transfer learning is a kind of deep neural network-based transfer learning method, which can transfer the knowledge learned from the source domain to the target domain through a set of labeled data. In general, transfer learning can be divided into two categories, i.e., inductive transfer learning and unsupervised transfer learning. • Inductive transfer learning: In this type of transfer learning, a neural network model is first trained with a large amount of unlabeled data, and then the neural network is fine-tuned with a small amount of data for a specific domain. After that, it is used to classify or predict the sentiment of a new target domain. The target domain contains a large number of different types of data, such as text, images, videos, and documents.

\paragraph{1) CNN-BASED MODELS}
Recently, CNN-based models have been widely used in NLP tasks, such as text classification BIB11 - BIB12, speech recognition , and optical image de-scattering. Compared with the traditional shallow neural network models, CNNs are able to capture the global structure of the input data. CNNs contain multiple convolutional layers, pooling layers, and fully connected layers, which can capture the local features in an end-to-end manner. Convolution and pooling operations in CNNs can be regarded as a kind of unsupervised feature learning method. In the field of NLP, CNN is one of the most successful models due to its ability to learn high-level abstractions from low-level image features. In recent years, CNN has achieved great success in computer vision and natural language processing (NLP) tasks, and has been successfully applied to many fields such as image classification, speech recognition, machine translation, and computer vision. However, due to the lack of transferability of CNNs to transfer learning tasks, most existing deep learning-based sentiment analysis models based on CNNs cannot be directly applied to sentiment analysis problems.

\paragraph{2) RNN-BASED MODELS}
RNN-based models are a type of neural networks that are used for processing sequential data. RNNs are a special type of deep neural networks, in which the hidden units of the neural network are connected to each other in a manner similar to the way that neurons in the brain memorize information. In other words, the hidden unit of a neural network is a vector or a tensor, and the output of the hidden layer is a binary value indicating the strength of the relation between the input and output. To make use of RNN in sentiment analysis, some researchers have applied RNNbased models to sentiment analysis tasks, such as BIB13 - BIB14, and BIB15. In general, the architecture of these models is shown in Fig. 4.

\paragraph{3) HYBRID NEURAL NETWORK MODELS}
The traditional neural network models, such as SVM, CNN, RNN, RBM, and RBM-NN, all have their pros and cons. Each model has its advantages and limitations. For instance, CNN has many layers of neurons, while RNN has only one layer of neurons. However, CNN and RNN have different levels of nonlinearity and therefore have different strengths and weaknesses. Combining these two kinds of neural networks can improve the performance of sentiment transfer learning models. In BIB16, the authors proposed a hierarchical attention network for text classification. The proposed model has a hierarchical structure that mirrors the hierarchical structure of documents and it has two levels of attention mechanisms applied at the word and sentence levels. The attention mechanism is applied to differentially attend differentially to more and less important content during the construction of the document representation. Experiments conducted on six large scale text classification tasks demonstrate that the proposed model outperforms previous methods by a substantial margin. BIB17 proposed a deep memory network for aspect level sentiment classification. This model explicitly captures the importance of each context word when inferring the sentiment polarity of an aspect. In this model, the importance degree and text representation are calculated with multiple computational layers, each of which is a

\subsection*{C. SENTIMENT ANALYSIS BASED ON TRANSFER LEARNING}
Transfer learning BIB04, BIB18 is a new branch of machine learning methods that aims to solve the problems where the training data and testing data are taken from the same or different domains. The difference between the source domain and the target domain is that the feature space and the data distribution characteristics of the source and target domains are the same. However, in some real-world situations, this assumption may not hold. Therefore, there are cases where training data are expensive or difficult to collect. There is a need to create high-performance learners trained with more easily obtained data from transfer learning. This methodology is referred to as transfer learning, which transfers knowledge from a source domain to a target domain BIB19. Transfer learning can be divided into two categories: inductive transfer learning and unsupervised transfer learning .

\paragraph{1) PARAMETER-TRANSFER METHODS}
These methods are based on the assumption that the source domain and target domain are drawn from the same distribution, which means that the probability distributions of the source and target data are similar. The parameter-transformation based transfer learning methods can be divided into two categories: Parameter-based and model-based. In the parameter-based method, the parameters of the target domain and source domain are learned simultaneously. In other words, the target and source domains are treated as two different domains, and the model parameters are jointly optimized to improve the performance of transfer learning. In this method, a deep neural network model is first trained on the source data, and then the parameters are used to initialize the target model. After that, the model is fine-tuned on the target data to achieve the best transfer learning performance. This method is also known as stacked denoising autoencoder (SDA) BIB20, BIB21, marginalized SDA BIB22, and Universal Language Model Fine-Tune (ULMFiT) method BIB23. 1) STacked Denoising Autoencoders (SDAs): In the SDA model, the encoder and decoder are trained simultaneously. The encoder learns the

\paragraph{2) INSTANCE-TRANSFER METHODS}
The instance-transfer methods aim to transfer knowledge from a source domain to a target domain by using a small amount of labeled training data and large amount of unlabeled data in the target domain. The advantage of these methods is that the source domain data can be used to improve the performance of the target-side classifier. However, these methods may face the problem that the distribution of training data is different between the source and target domains. Therefore, it is necessary to find a balance between the training data distribution and the target distribution in order to achieve good transfer learning performance. The Instance-Transfer methods can be divided into two categories: instance-based methods and instance-tracing methods. In the following, we will introduce these methods in detail. The first method is the instancebased method. In this method, the target data is firstly transformed into the source data, and then the target samples are used to train the target classifier by using the labeled target data. The second method is to use the labeled source data to initialize the training process of the new target domain, which is called the transfer learning method. For instance, in BIB24, a novel transfer learning framework called TrAdaBoost is proposed, which extends boosting-based

\paragraph{3) FEATURE-REPRESENTATION-TRANSFER METHODS}
In order to bridge the gap between domains, feature-representation-based transfer methods can be used to transfer knowledge from the source domain to the target domain. In this kind of methods, the feature transformation matrix of the source and target domain can be obtained by mapping the feature space into a common latent space. Then, the classifiers trained on the source data can be easily applied to solve the target problem by using the common space BIB25, BIB26. Feature representation-based methods mainly include co-clustering methods, semi-supervised methods, and inductive transfer learning methods. The co- Clustering method is to find the co-occurrence patterns of words in the common latent representation space, and then the domain-independent words are used as the bridge between domains. In this method, the similarity between the feature spaces of the two domains is measured by computing the distance between the two feature spaces. The similarity measure can be based on the cosine similarity, the Kullback-Leibler divergence, or the Jaccard similarity. The Co-Clustering-based Transfer Learning (CCL) method is a semisupervised method that uses unlabeled target data and labeled source data

\paragraph{4) SUMMARY}
In this paper, we summarize the current state-of-the-art of transfer-based sentiment classification methods for sentiment transfer learning in the following three aspects: (1) classification methods, (2) clustering methods, and (3) fine-grained transfer learning methods. Classification methods are mainly divided into two categories: supervised learning and unsupervised learning. Clustering methods are divided into co-clustering and hierarchical clustering. Hierarchical clustering is mainly used for dimensionality reduction and feature representation learning. Feature representation learning is used to bridge the distribution gap between different feature spaces. Transfer learning is divided into inductive and inductive transfer learning. Inductive transfer learning focuses on transferring knowledge from a source domain to a target domain. On the other hand, transfer learning can be divided into semi-supervised and supervised learning. Supervised learning is based on labeled data in the source domain and unlabeled data only in the target domain, and transfer learning is to transfer knowledge between domains. In addition, the methods based on co-occurrence matrix and spectral feature alignment (SFA) BIB27, TCT BIB28, ULMFiT, LSTM-CNN

\subsection*{A. CROSS-DOMAIN SENTIMENT ANALYSIS}
In order to solve the problem of cross-domain sentiment analysis, transfer learning has been widely used in recent years BIB26, BIB28, . The main idea of transfer learning is to build a bridge between the source domain and the target domain by transferring knowledge from a source domain to a target domain BIB15 - BIB12. Cross-domain transfer learning based sentiment analysis can be divided into two categories: (1) unsupervised transfer learning and (2) supervised transfer learning. In the following, we will introduce the two sub-tasks in more detail.

\section*{References}

\begin{itemize}
\setlength{\leftskip}{-0.4cm}
    
\item[] \textbf{BIB01} Mohit Mertiya and Ashima Singh. (2016). Combining naive bayes and adjective analysis for sentiment detection on Twitter.
    
\item[] \textbf{BIB02} Erik Cambria. (2016). Affective Computing and Sentiment Analysis.
    
\item[] \textbf{BIB03} María del Pilar Salas-Zárate, José Medina-Moreira, Paul Javier Álvarez-Sagubay, Katty Lagos-Ortiz, Mario Andrés Paredes-Valverde, and Rafael Valencia-García. (2016). Sentiment Analysis and Trend Detection in Twitter.
    
\item[] \textbf{BIB04} Sinno Jialin Pan and Qiang Yang. (2010). A Survey on Transfer Learning.
    
\item[] \textbf{BIB05} E. Cambria, B. Schuller, Yunqing Xia, and C. Havasi. (2013). New Avenues in Opinion Mining and Sentiment Analysis.
    
\item[] \textbf{BIB06} Rui Xia, Feng Xu, Chengqing Zong, Qianmu Li, Yong Qi, and Tao Li. (2015). Dual Sentiment Analysis: Considering Two Sides of One Review.
    
\item[] \textbf{BIB07} Peter D. Turney and Michael L. Littman. (2003). Measuring praise and criticism: Inference of semantic orientation from association.
    
\item[] \textbf{BIB08} Xiaoxu Fei, Huizhen Wang, and Jingbo Zhu. (2010). Sentiment word identification using the maximum entropy model.
    
\item[] \textbf{BIB09} Huimin Lu, Yujie Li, Min Chen, Hyoungseop Kim, and Seiichi Serikawa. (2017). Brain Intelligence: Go Beyond Artificial Intelligence.
    
\item[] \textbf{BIB10} Huimin Lu, Yujie Li, Shenglin Mu, Dong Wang, Hyoungseop Kim, and Seiichi Serikawa. (2018). Motor Anomaly Detection for Unmanned Aerial Vehicles Using Reinforcement Learning.
    
\item[] \textbf{BIB11} Rie Johnson and Tong Zhang. (2014). Effective Use of Word Order for Text Categorization with Convolutional Neural Networks.
    
\item[] \textbf{BIB12} Alexis Conneau, Holger Schwenk, Loïc Barrault, and Yann LeCun. (2017). Very Deep Convolutional Networks for Text Classification.
    
\item[] \textbf{BIB13} Duyu Tang, Bing Qin, and Ting Liu. (2015). Document Modeling with Gated Recurrent Neural Network for Sentiment Classification.
    
\item[] \textbf{BIB14} Min-Yuh Day and Yue-Da Lin. (2017). Deep Learning for Sentiment Analysis on Google Play Consumer Review.
    
\item[] \textbf{BIB15} Rie Johnson and Tong Zhang. (2016). Supervised and Semi-Supervised Text Categorization using LSTM for Region Embeddings.
    
\item[] \textbf{BIB16} Zichao Yang, Diyi Yang, Chris Dyer, Xiaodong He, Alexander J. Smola, and Eduard H. Hovy. (2016). Hierarchical Attention Networks for Document Classification.
    
\item[] \textbf{BIB17} Duyu Tang, Bing Qin, and Ting Liu. (2016). Aspect Level Sentiment Classification with Deep Memory Network.
    
\item[] \textbf{BIB18} Karl Weiss, Taghi M. Khoshgoftaar, and DingDing Wang. (2016). A survey of transfer learning.
    
\item[] \textbf{BIB19} Diane Cook, Kyle D. Feuz, and Narayanan C. Krishnan. (2013). Transfer learning for activity recognition: a survey.
    
\item[] \textbf{BIB20} Xavier Glorot, Antoine Bordes, and Yoshua Bengio. (2011). Domain Adaptation for Large-Scale Sentiment Classification: A Deep Learning Approach.
    
\item[] \textbf{BIB21} Minmin Chen, Zhixiang Xu, Kilian Weinberger, and Fei Sha. (2012). Marginalized Denoising Autoencoders for Domain Adaptation.
    
\item[] \textbf{BIB22} Miao Sun, Qi Tan, Runwei Ding, and Hong Liu. (2014). Cross-domain sentiment classification using deep learning approach.
    
\item[] \textbf{BIB23} Jeremy Howard and Sebastian Ruder. (2018). Universal Language Model Fine-tuning for Text Classification.
    
\item[] \textbf{BIB24} Wenyuan Dai, Qiang Yang, Gui-Rong Xue, and Yong Yu. (2007). Boosting for transfer learning.
    
\item[] \textbf{BIB25} John Blitzer, Ryan McDonald, and Fernando Pereira. (2006). Domain Adaptation With Structural Correspondence Learning.
    
\item[] \textbf{BIB26} Sinno Jialin Pan, Xiaochuan Ni, Jian-Tao Sun, Qiang Yang, and Zheng Chen. (2010). Cross-domain sentiment classification via spectral feature alignment.
    
\item[] \textbf{BIB27} Joey Tianyi Zhou, Ivor W. Tsang, Sinno Jialin Pan, and Mingkui Tan. (2014). Heterogeneous Domain Adaptation for Multiple Classes.
    
\item[] \textbf{BIB28} Guangyou Zhou, Yin Zhou, Xiyue Guo, Xinhui Tu, and Tingting He. (2015). Cross-domain sentiment classification via topical correspondence transfer.

\end{itemize}

\label{fullex2}

\end{document}